%% file: paper1850.tex
\documentclass[runningheads]{llncs}
\usepackage{graphicx} 
\usepackage[detect-all]{siunitx}
\usepackage[usenames, dvipsnames]{color} 
\usepackage{makecell}
\usepackage{hyperref}
\usepackage{multirow}  
\input{tex/defs}

\begin{document}
%
%\title{Incorporating prior knowledge by Orientation Estimation of Structures in Volumetric Data}
\title{Probabilistic Atlases to Enforce Topological Constraints}

\author{Udaranga Wickramasinghe\inst{1} \and
Graham Knott\inst{2} \and
Pascal Fua\inst{1}}
% index{Wickramasinghe, Udaranga}
% index{Knott, Graham}
% index{Fua, Pascal}
 
\authorrunning{U. Wickramasinghe et al.}
% First names are abbreviated in the running head.
% If there are more than two authors, 'et al.' is used.
%
\institute{
Computer Vision Laboratory, \'Ecole Polytechnique F\'ed\'erale de Lausanne, Switzerland \\
\email{udaranga.wickramasinghe@epfl.ch}
\and BioEM Laboratory, 
\'Ecole Polytechnique F\'ed\'erale de Lausanne, Switzerland}

\maketitle              % typeset the header of the contribution
%
 
%
%
\input{tex/0_abstract.tex}
\input{tex/1_introduction.tex}

\input{tex/2_related_work.tex}

\input{tex/3_method.tex}

\input{tex/4_experiments.tex}
\input{tex/5_conclussion.tex}
\input{tex/6_acknowledgment.tex}

\bibliographystyle{splncs04}
\bibliography{biomed,vision}

\end{document}

%% file: tex/defs.tex
\newif\ifdraft
\draftfalse
\definecolor{orange}{rgb}{1,0.5,0}
\definecolor{violet}{RGB}{70,0,170}
\definecolor{magenta}{RGB}{170,0,170}
\definecolor{dgreen}{RGB}{0,150,0}

\newcommand{\mA}{\mathcal{A}}

\newcommand{\bX}{\mathbf{X}}
\newcommand{\bY}{\mathbf{Y}}

\newcommand{\mT}{\mathcal{T}}
\newcommand{\mL}{\mathcal{L}}
\newcommand{\mF}{\mathcal{F}}

\newcommand{\bq}{\mathbf{q}}

\newcommand{\mD}{\mathcal{D}}
\newcommand{\mE}{\mathcal{E}}

\newcommand{\panet}[0]{\textbf{PA-Net}}

%% file: tex/0_abstract.tex
% !TEX root = ../top.tex
% !TEX spellcheck = en-US

\begin{abstract}

Probabilistic atlases (PAs) have long been used in standard segmentation approaches and, more recently, in conjunction with Convol-utional Neural Networks (CNNs). However, their use has been restricted to relatively standardized structures such as the brain or heart which have limited or predictable range of deformations. Here we propose an encoding-decoding CNN architecture that can exploit rough atlases that encode only the topology of the target structures that can appear in any pose and have arbitrarily complex shapes to improve the segmentation results.  It relies on the output of the encoder to compute both the pose parameters used to deform the atlas and the segmentation mask itself, which makes it effective and end-to-end trainable. 

\end{abstract}

%% file: tex/1_introduction.tex
% !TEX root = ../top.tex
% !TEX spellcheck = en-US

\section{Introduction}

Probabilistic atlases (PAs) are widely used for multi-atlas based segmentation\cite{Eugenio15}. With the advent of deep learning, there has been a push to incorporate them with  Convolutional Neural Networks (CNNs) as well \cite{Atzeni18,Yuankai18,Spitzer17,Hannah18}. The published techniques relying on deep learning share a number of features: They work best for structures featuring relatively small variations in shape and position; the atlases are often  created by fusing multiple manually annotated images;  the atlases must also be pre-registered to the target images to align them with the structures of interest.  

Thus, the PAs have been mostly used to segment structures such as the brain  or heart for which the above requirements can be met. In this paper, we propose an approach to design and handle PAs that makes them usable in complex situations where the shape and position can vary dramatically, such as those shown in Fig.~\ref{fig:teaser}. Our PAs are coarse and only encode the relative position and topology of the structures we expect to find.  To register them, we rely on affine transforms whose parameters are estimated at the same time as the segmentations themselves using the end-to-end trainable encoder-decoder architecture depicted by Fig.~\ref{fig:arch}, which we will refer to as \panet{}. The affine transforms are used to warp the atlases and feed the features of the warped atlases to the decoder.  This differs significantly from earlier approaches~\cite{Spitzer17,Hannah18} that rely on pre-registered atlases.

\input{fig/teaser}
We validate our approach on segmentation of synaptic junctions in Electron Microscopy (EM) images and optic nerve head segmentation in retinal fundus images. A synaptic junction has an arbitrary shape but always features a synaptic cleft sandwiched between a pre-synaptic bouton belonging to the axon of one neuron and a post-synaptic dendritic spine belonging to another. In this case, the shape complexity and unpredictability precludes the use of standard PAs. Optic nerve head consists of an optic disk and optic cup. Even though their shape does not vary significantly, there are significant variations in the size of the optic cup and the position of the optic nerve head. 
 
We will show that \panet{} eliminates most of the topological mistakes its unaided V-Net~\cite{Milletari16} or U-Net~\cite{Ronneberger15} backbone makes on the EM and retinal images, while being generic and potentially applicable to other segmentation tasks. Our contribution is therefore a novel approach that makes it possible to use probabili- -stic atlases even in situations where shape variability is too great for existing approaches and methods to integrate the atlas registration process directly into the network. The code is publicly available\footnote{\url{https://github.com/cvlab-epfl/PA-net.git} }.

\input{fig/arch}

%% file: fig/teaser.tex
% !TEX root = ../top.tex
% !TEX spellcheck = en-US

\begin{figure}
\centering
\begin{tabular}{cccc}
    \includegraphics[width=0.2\textwidth]{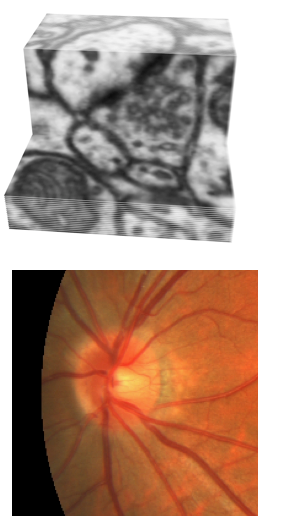}&
    \includegraphics[width=0.2\textwidth]{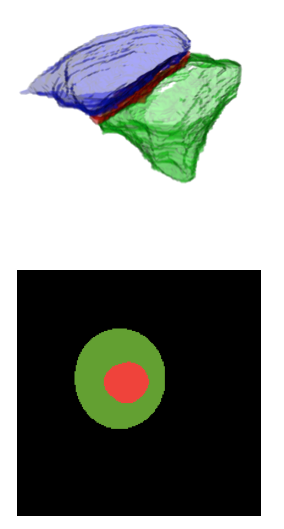}&
     \includegraphics[width=0.2\textwidth]{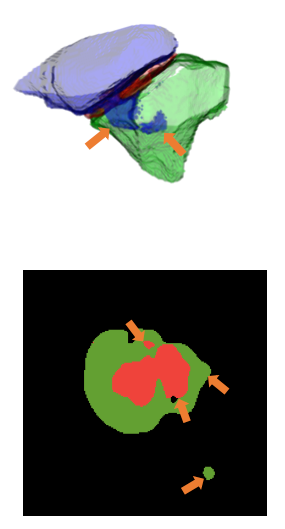}&
      \includegraphics[width=0.2\textwidth]{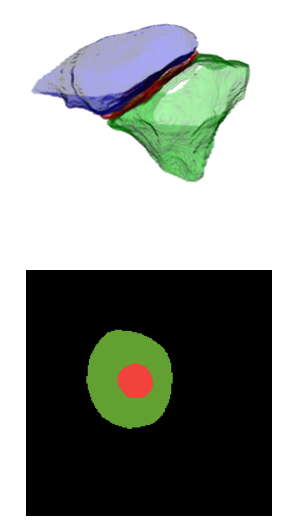}\\
      (a)&(b)&(c)&(d)
\end{tabular}    
\caption{Eliminating topological mistakes. (a) A small FIBSEM image stack featuring a synaptic junction and a retinal fundus image. (b) Ground-truth segmentations. The pre-synaptic, synaptic-cleft, and post-synaptic regions shown in green, red, and blue respectively. Similarly, the optic disk and cup shown in green and red, respectively. (c) U-Net segmentations with orange arrows pointing to topological  mistakes. (d) Our error-free result (best viewed in color).}
\label{fig:teaser}
\end{figure}

%% file: fig/arch.tex
% !TEX root = ../top.tex
% !TEX spellcheck = en-US

\begin{figure}
\centering
\includegraphics[height=4.5cm]{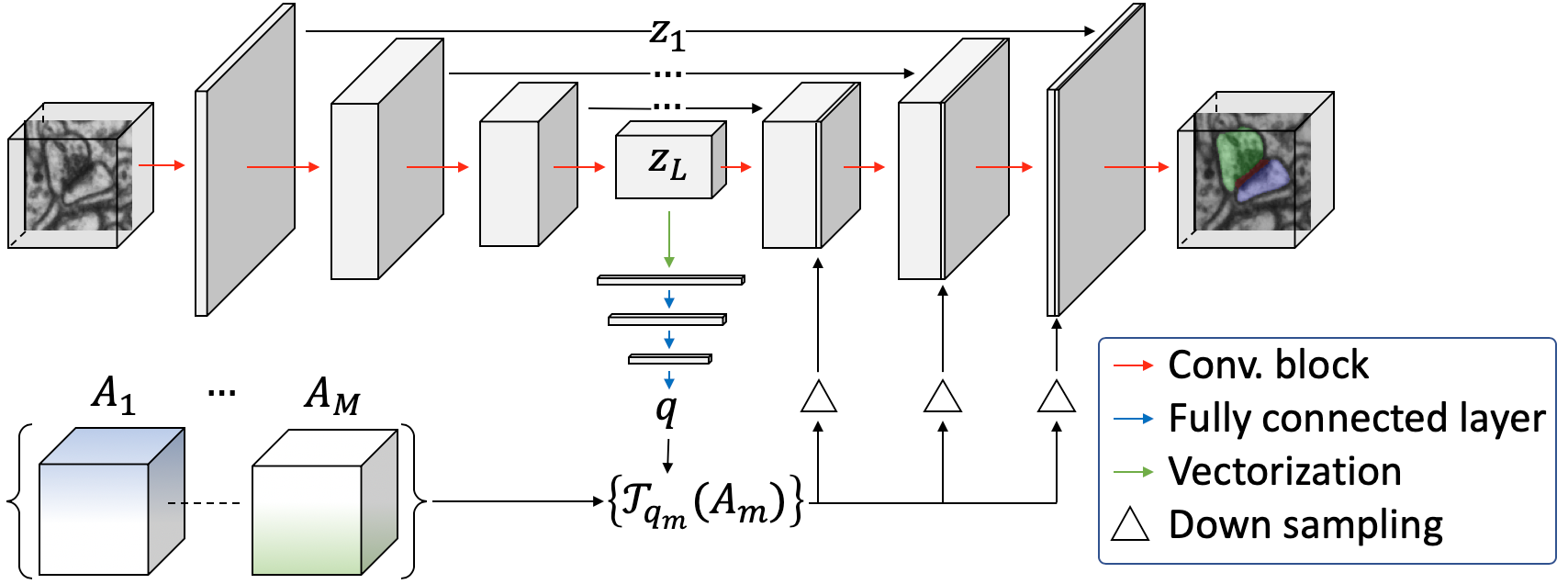}
\caption{\panet{} architecture. The network takes as input an image and a list of PAs. Given the decoder output, the fully connected layers estimate the parameters of the affine transformation that registers the PAs to the input. The registered PAs are then concatenated with the features computed by the convolutional decoder at each scale. }
\label{fig:arch}
\end{figure}

%% file: tex/2_related_work.tex
% !TEX root = ../top.tex
% !TEX spellcheck = en-US

\section{Related Work} 

PAs are commonly used in standard atlas-based segmentation approaches. In their piplines, CNNs are indirectly used to perform segmentation. These include using a CNN to infer deformation maps to register target images to atlas images \cite{Balakrishnan18} and using it to implicitly infer the segmentation masks\cite{Dong18,Alessa19};  using CNNs to improve the selection of best atlases from a library~\cite{Katouzian18}. As this is not the focus of this paper, we will not discuss them further and will concentrate on methods that directly use the PAs to assist a CNN that perform segmentation.

A  PA can provide localization priors to help a network find the target of interest. Such PAs are often referred to as seed layers. They come in two flavors,  Gaussian priors~\cite{Bertasius17,Yang18e}  or binary seed layers~\cite{Januszewski18}. Another popular way to focus the attention of the network is to provide an attention mask~\cite{Fei17}, which then serves the same purpose as a PA. However, the attention masks are derived from the input data itself and does not act as an external knowledge source.

PAs built by fusing multiple manually annotated images can be used to introduce even more prior knowledge about shape and topology of structures. In~\cite{Spitzer17,Hannah18}, this is done to improve segmentations of human brain by having the CNN take pre-registered PAs as input. In our work, we extend the idea by demonstrating that PAs can be used to introduce topological knowledge even when the structure of interest exhibits large shape variations and without pre-registration that the whole system is end-to-end trainable.

%% file: tex/3_method.tex
% !TEX root = ../top.tex
% !TEX spellcheck = en-US
 
\section{Approach}

\subsection{Network Architecture}
The architecture of the proposed \panet{} is depicted in Fig.~\ref{fig:arch}. Its backbone is an encoder-decoder architecture similar to the one used in popular networks such as U-Net~\cite{Ronneberger15} or V-Net~\cite{Milletari16}. These networks are designed to learn to approximate the distribution $p( \bY | \bX)$, where $\bX$ is the input image and $\bY$ is the output segmentation. We extend the idea by learning the joint distribution $ p( \bY, T_{\bq}(\mA)  | \bX, \mA)$. The function $\mT_{\bq}(\cdot)$ applies affine transformations to the set of PAs $\mA$ given the pose vectors $\bq$. We factorize this joint distribution as 
 \begin{equation}
 p( \bY, T_{\bq}(\mA) | \bX, \mA) = p(\bY | \bX, T_{\bq}(\mA))p(\bq | \bX, \mA)
 \label{eq:jointdist}
\end{equation}
\panet{} models it using a primary stream, the encoder-decoder backbone at the top of Fig. \ref{fig:arch}, and a secondary stream, the fully connected layers at the bottom of Fig. \ref{fig:arch}. The secondary stream $\mF(\cdot,\theta_f)$ originates from the latent vector produced by the final stage of the encoder $\mE(\cdot,\theta_e)$, estimates the affine parameter vector $\bq$, uses it to warp the atlases, and feed them to the various layers of the decoder $\mD(\cdot;\theta_d)$ after rescaling them to the resolution of feature vectors at each stage. Here  $\theta_e, \theta_d$ and $\theta_f$ are the learned weights that control the behavior of $\mE, \mD$ and $\mF$.  Given the encoder output,  $\mF$ and $\mD$ model $p(\bY | \bX, T_{\bq}(A))$  and $p(\bq | \bX, A)$, respectively.  This guarantees that these two distributions are learned from the same features. Training the encoder to produce features that are effective for both tasks yield improved performance.   

\subsection{Atlas Design} 

Fig \ref{fig:PAs}. depicts the topological atlases we use to segment synaptic junctions and optic disks and cups. In the first case, they are cubes and in the second, ordinary 2D arrays. They encode the probability of any pixel of voxel to belong to one of the possible classes---pre-synaptic, cleft, or post-synaptic for synapses and optic disk or cup for retinas---given its pose. 

\input{fig/PA}

\subsection{Atlas Registration}

Parameters necessary for PA registration is computed by estimating the pose vector $q$. When working with image cubes such as the one shown at the top of Fig.~\ref{fig:teaser}, we take the pose vector to be $\bq =[t_x,t_y,t_z, s_x, s_y, s_z, r_x,r_y,r_z]$, where $t_{\{x,y,z\}}$ represent translations and  $s_{\{x,y,z\}}$ scalings in the $x$,$y$, and $z$ directions, respectively. We represent the orientation using the angle-axis vector $[r_x,r_y,r_z]$, whose direction is the axis of rotation. When working with 2D images,  we drop $t_z$, $s_z$, and $r_z$ from $\bq$. 

\input{tables/synapses} 

\subsection{Loss Functions}
 
Following standard practice, we formulate the loss terms as the negative log likelihood of the joint distribution $p( \bY, T_{\bq}(A) | \bX, \mA) $ of Eq.~\ref{eq:jointdist}. This yields the composite loss
 \begin{equation}
 \mL = \mL_{seg}  + \mL_{pose} \; .
\end{equation}  
$\mL_{seg}$ is the standard cross entropy loss  that evaluates segmentation performance and take $\mL_{pose}$ is the least square error in estimating the pose $\bq$.

%% file: fig/PA.tex
% !TEX root = ../top.tex
% !TEX spellcheck = en-US

\begin{figure}
\centering
\begin{tabular}{ccc}
    \includegraphics[width=0.45\textwidth]{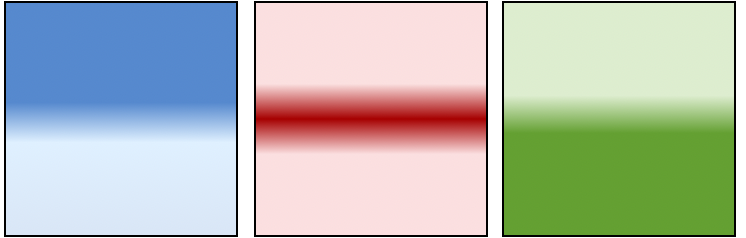}&~~~~&
    \includegraphics[width=0.3\textwidth]{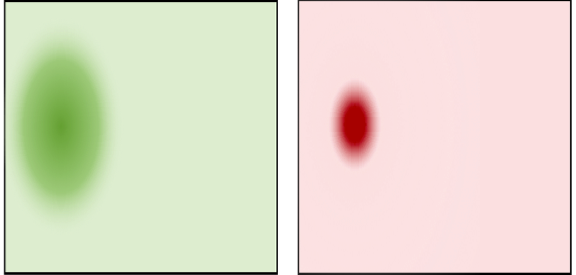}\\
      (a)& ~ &(b) 
\end{tabular}    
\caption{Probabilistic atlases. (a) For synaptic junction segmentation, they are cubes and we show a single face. (b) For optic optic disk and cup segmentation, they are regular 2D arrays. The color hues denote the same areas as those in Fig. \ref{fig:teaser}. The color saturation is proportional to the probability at a given location (best viewed in color). }
\label{fig:PAs}
\end{figure}

%% file: tables/synapses.tex
\begin{table}
\centering
\caption{Comparative results for synaptic junction segmentation.}
\label{table:synapses}
\begin{tabular}{|l|c|c|c|c|c|}
\hline
\multirow{2}{*}{~} & \multicolumn{4}{c|}{Jaccard index (\%)} & \multirow{2}{*}{TER} \\
\cline{2-5}
 & Pre-synaptic & Synaptic junction & Post-synaptic & Mean  &  \\ 
\hline 
V-Net & 		\textbf{63.0} & 				56.0 & 			  80.6 & 			 66.5 & 5/15 \\ 
\hline 
PA-VNet & 				59.5 & \textbf{57.8} & \textbf{83.1} & \textbf{66.8} &  \textbf{1/15} \\  
\Xhline{3\arrayrulewidth}
U-Net & 	    \textbf{73.6} & 	 			59.7 & 				79.1 & 			   70.7  &  7/15 \\ 
\hline 
Naive PA-UNet & 	73.4 &  58.0 & 78.4 & 69.9 & 7/15\\  
\hline 
PA-UNet & 				72.6 &  \textbf{62.8} & \textbf{87.0} & \textbf{74.1} & \textbf{3/15}\\  
\Xhline{3\arrayrulewidth}
\end{tabular}
\end{table}

%% file: tex/4_experiments.tex
% !TEX root = ../top.tex
% !TEX spellcheck = en-US

\section{Results and Discussion}

\subsection{Datasets}

\subsubsection{Synaptic Junction Dataset:} It is a $500 \times 500 \times 200$ FIB-SEM image stack of a mouse cortex. We used 50 xy slices for training, 50 for validation, and 100 for testing. From each set, we cropped $96\times96\times96$ image volumes containing a synaptic junction such as the one shown in the top row of Fig. \ref{fig:teaser}(a) and they are zero-padded as necessary. This gave us 13, 10 and 15 volumes for training, validation and testing, respectively. 
The synapse is not necessarily perfectly centered and the task is to segment pre-synaptic region, post-synaptic region, and synaptic cleft. 

\subsubsection{Retinal Fundus Image Dataset:} It comprises 400 $2124 \times 2056$ retinal fundus images acquired using a Zeiss Visucam 500 camera \cite{Rfgc18}, which we resize and pad to be $512 \times 512$. We use 100 for training, 100 for validation, and 200 for testing. 
\input{tables/retina} 
\subsection{Quantitative and Qualitative Results}

We evaluate segmentation performance in terms of the standard Jaccard Index and of an additional metric we dub the \textit{Topological Error Ratio}(TER). We define  TER as the ratio of segmentations containing  topological errors to the total number of test images. A segmentation is considered to contain a topological error if it violates the expected topology of the target structure, that is, it features semantic labels appearing where they should not given the topological constraints. We compute this value by finding the connected components of the final segmentation and then finding instances that violate the topology.

We tested two versions of \panet{}, one based on the U-Net~\cite{Ronneberger15} architecture and the other based on the more recent V-Net~\cite{Milletari16}. We compare the results against those of the standard U-Net and V-Net. We report our comparative results on our two datasets in Tables~\ref{table:synapses} and ~\ref{table:retina}. Figs.~\ref{fig:results_all}  depict them qualitatively (see supplementary materials for further results). Using the atlas consistently improves over baseline performance. The gains are most visible in TER terms because our \panet{}s truly come into their own when the standard U-Net and V-Net fail, which is only a fraction of the time, as would be expected of architectures that are as popular as they are. Even when it fails to eliminate topological errors, it reduces the size of the errors as shown in the second row of Fig. \ref{fig:results_all}). The one exception is the pre-synaptic region, for which the Jaccard numbers decrease. A closer inspection of the results show that the atlas sometimes encourages the segmentation to leak from weak boundary regions in few instances. We plan to address this issue by introducing  an additional stream that specifically focus on predicting object boundaries and combining its results with \panet{}.

To demonstrate the importance of using the same features to compute the segmentation {\it and} the pose parameters, we implemented a naive version of our approach that uses two {\it separate} streams to compute the pose and the segmentati- -on features, which we denote as  Naive PA-Unet in Tables~\ref{table:synapses} and~\ref{table:retina}. The naive version fails to resolve the topological mistakes in the datasets. In the synaptic junction dataset, this occurs because the naive version has a mean orientation estimation error of \ang{80.6} in contrast to the \panet{}'s error of \ang{10.4}. In the retinal fundus image dataset, the naive version has a mean localization error of $26.5$ pixels compared to a $3.9$ pixel error for \panet{}. As a result, in both instances, network fails to properly register PAs.
\input{fig/results_all}

%% file: tables/retina.tex
\begin{table}
\centering
\caption{Comparative results for  fetinal fundus image segmentation.}
\label{table:retina}
\begin{tabular}{|l|c|c|c|c|c|}
\hline
\multirow{2}{*}{~} & \multicolumn{3}{c|}{Jaccard index (\%)} & \multirow{2}{*}{TER} \\
\cline{2-4}
 & Optic disk & Optic cup  & Mean & \\ 
\hline 
V-Net & 78.4 & 72.5  & 75.5 & 7/200 \\ 
\hline 
PA-VNet & \textbf{79.0} &  \textbf{73.0} &\textbf{ 76.0} & 3/200 \\ 
\Xhline{3\arrayrulewidth}
U-Net & 81.3 & 73.8  & 77.6 & 18/200 \\ 
\hline 
Naive PA-UNet & 80.8 & 74.1  & 77.3 & 16/200 \\ 
\hline 
PA-UNet & \textbf{81.6} &  \textbf{74.6} & \textbf{78.1} & \textbf{5/200} \\ 
\Xhline{3\arrayrulewidth}
\end{tabular} 
\end{table}

%% file: fig/results_all.tex
% !TEX root = ../top.tex
% !TEX spellcheck = en-US
\begin{figure}
 \centering
\includegraphics[width=10.6cm]{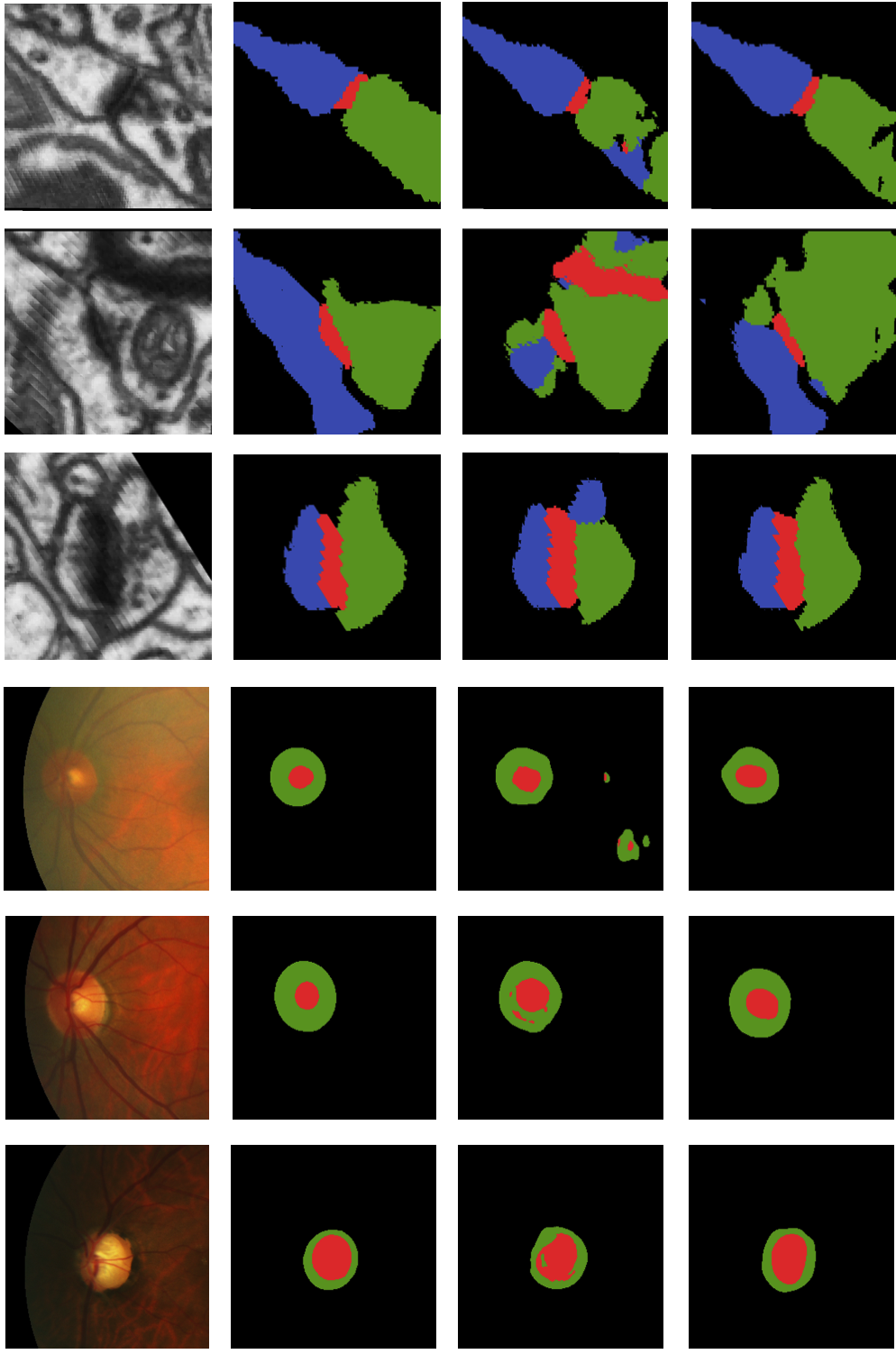}\\
(a)\hspace{2.3cm}(b)\hspace{2.3cm}(c)\hspace{2.3cm}(d)
\caption{U-Net vs \panet{}. Top three rows depict results from synaptic junction segmentation. The bottom three rows depict results from retinal fundus image segmentation. (a) Input images, (b)  Ground-truth, (c) U-Net results (d) \panet{} results. The color hues denote the same areas as those in Fig. \ref{fig:teaser} (best viewed in color).  }
\label{fig:results_all}
\end{figure}

%% file: tex/5_conclussion.tex
% !TEX root = ../top.tex
% !TEX spellcheck = en-US

\section{Conclusion}

We have proposed an encoding-decoding architecture that can exploit rough atlases that encode the topology of structures that can appear in any pose and have arbitrarily complex shapes to improve the segmentation results.  One of its crucial components is that it relies on the output of the encoder to compute both the pose parameters used to deform the atlas and the segmentation mask itself, which makes it effective and end-to-end trainable. As future work, we plan to extend \panet{} by introducing an additional stream to produce edge maps and using it to address the minor loss in accaracy that occur when the structures have weak boundaries. 

%% file: tex/6_acknowledgment.tex
% !TEX root = ../top.tex
% !TEX spellcheck = en-US
\subsubsection{Acknowledgments:} This work was supported in part by a Swiss National Science Foundation grant.